\documentclass[lettersize,journal]{IEEEtran}
\usepackage{amsmath,amsfonts}
\usepackage{algorithmic}
\usepackage{algorithm}
\usepackage{array}
\usepackage[caption=false,font=normalsize,labelfont=sf,textfont=sf]{subfig}
\usepackage{textcomp}
\usepackage{stfloats}
\usepackage{url}
\usepackage{verbatim}
\usepackage{graphicx}
\usepackage{cite}
\usepackage{amssymb}
\usepackage{multirow}
\usepackage[dvipsnames]{xcolor}
\newcommand{\eat}[1]{}
\usepackage{wrapfig}
\usepackage{float}
\usepackage{multicol}
\usepackage{cuted}
\usepackage{stfloats}
\usepackage{caption}
\usepackage[pagebackref=true,breaklinks=true,letterpaper=true,colorlinks,bookmarks=false]{hyperref}
\begin{document}

\title{Scale-Aware Pre-Training for Human-Centric Visual Perception: \\ Enabling Lightweight and Generalizable Models}

\author{Xuanhan Wang,
	Huimin Deng,
	Lianli Gao,~\IEEEmembership{Member,~IEEE,} \\
	and Jingkuan Song,~\IEEEmembership{Senior Member,~IEEE}
	
	\thanks{Xuanhan Wang is with Shenzhen Institute for Advanced Study, University of Electronic Science and Technology of China, Shenzhen, China (e-mail: wxuanhan@hotmail.com).}
	\thanks{Huimin Deng and Lianli Gao are with the Center for Future Media, University of Electronic Science and Technology of China, Chengdu, China (e-mail: hmdeng@std.uestc.edu.cn; lianli.gao@uestc.edu.cn).}
	\thanks{Jingkuan Song is with TongJi University, Shanghai, China (e-mail: jingkuan.song@gmail.com).}
}

\markboth{Journal of \LaTeX\ Class Files,~Vol.~14, No.~8, August~2021}%
{Shell \MakeLowercase{\textit{et al.}}: A Sample Article Using IEEEtran.cls for IEEE Journals}


\maketitle

\begin{abstract}
	Human-centric visual perception (HVP) has recently achieved remarkable progress due to advancements in large-scale self-supervised pretraining (SSP). However, existing HVP models face limitations in adapting to real-world applications, which require general visual patterns for downstream tasks while maintaining computationally sustainable costs to ensure compatibility with edge devices. These limitations primarily arise from two issues: 1) the pretraining objectives focus solely on specific visual patterns, limiting the generalizability of the learned patterns for diverse downstream tasks; and 2) HVP models often exhibit excessively large model sizes, making them incompatible with real-world applications.
	To address these limitations, we introduce Scale-Aware Image Pretraining (SAIP), a novel SSP framework pretraining lightweight vision models to acquire general patterns for HVP. Specifically, SAIP incorporates three learning objectives based on the principle of cross-scale consistency: 1) Cross-scale Matching (CSM) which contrastively learns image-level invariant patterns from multi-scale single-person images; 2) Cross-scale Reconstruction (CSR) which learns pixel-level consistent visual structures from multi-scale masked single-person images; and 3) Cross-scale Search (CSS) which learns to capture diverse patterns from multi-scale multi-person images. Three objectives complement one another, enabling lightweight models to learn multi-scale generalizable patterns essential for HVP downstream tasks.
	Extensive experiments conducted across 12 HVP datasets demonstrate that SAIP exhibits remarkable generalization capabilities across 9 human-centric vision tasks. Moreover, it achieves significant performance improvements over existing methods, with gains of 3\%-13\% in single-person discrimination tasks, 1\%–11\% in dense prediction tasks, and 1\%–6\% in multi-person visual understanding tasks. 
	Code and models are released\footnote{\url{https://github.com/stoa-xh91/SAIPv1}} for research purpose.
\end{abstract}


\section{Introduction}

\begin{figure}[ht]
	\centering
	\includegraphics[width=0.99\linewidth]{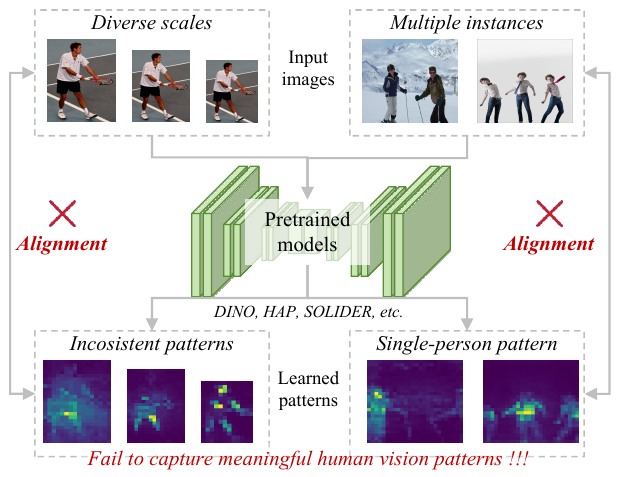}
	\caption{HVP models pretrained through existing SSP struggle to capture meaningful human visual patterns, especially in real-world scenarios that involve diverse scales or multiple instances. }
	\vspace*{-0.1in}
	\label{fig:domain_change}
\end{figure}

\begin{figure*}[ht]
	\hsize=\textwidth
	\centering
	\includegraphics[width=0.99\textwidth]{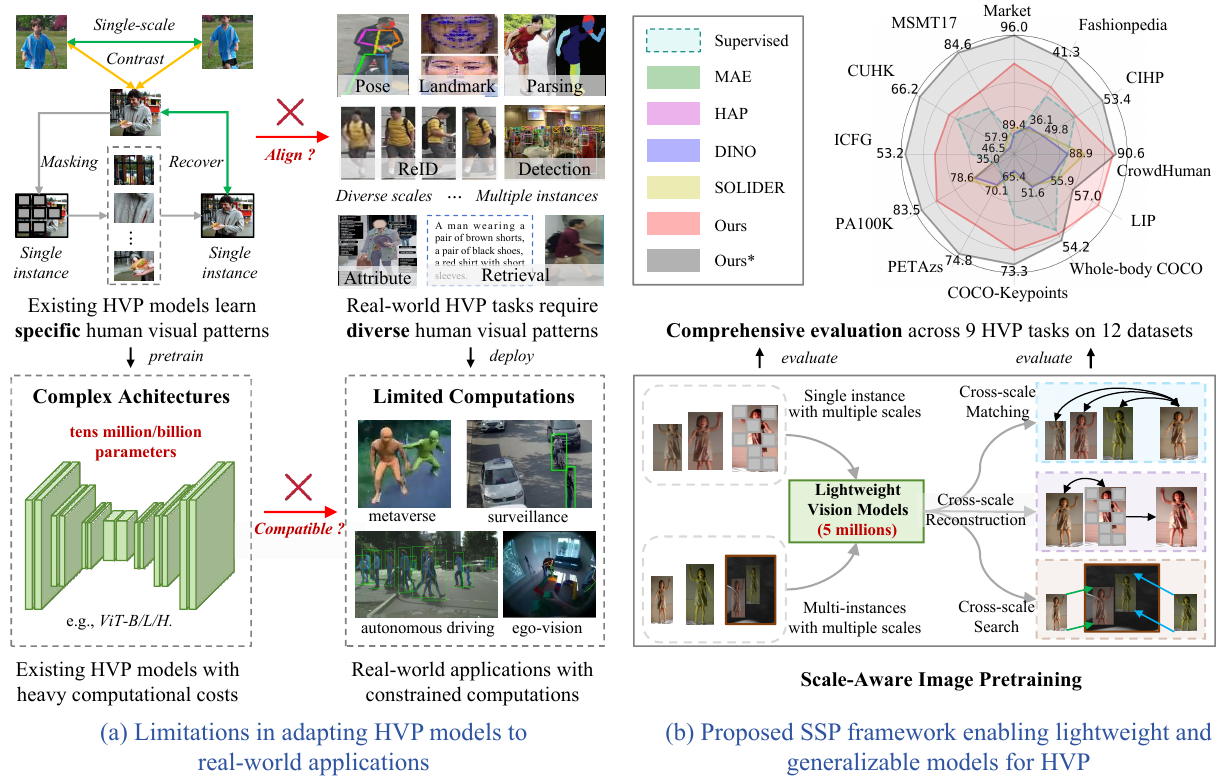}
	\caption{The motivation. (a) Human-centric vision models face limitations in adapting to real-world applications: 1) visual patterns learned through existing SSP fail to align with the general patterns required for real-world HVP tasks; and 2) the complex architectures of such models are incompatible with edge devices. (b) To address these limitations, Scale-Aware Image Pretraining (SAIP) is proposed to enable lightweight vision models to learn general visual patterns. Comprehensive evaluations across a wide range of HVP tasks demonstrate the effectiveness of proposed method. * denotes the enhanced version of SAIP. Zoom in for a better view.}
	\vspace*{-1.em}
	\label{figure:introduction}
\end{figure*}

With the advances in large-scale self-supervised pretraining (SPP)~\cite{ssp_dino,ssp_mae,ssp_eva,ssp_mae-lite,ssp_mae_dino,ssp_maefuse, ssp_tinymim}, remarkable progress has been made in human-centric visual perception (HVP). Numerous general HVP models \cite{hc_ssp_hap,hc_ssp_hcmoco,hc_ssp_humanbench,hc_ssp_solider,hc_ssp_unihcp} have been introduced, substantially advancing the state of the art. 
Despite the differences in architectures, existing general HVP models follow a common training paradigm: they utilize self-supervised pretraining to enable vision models to learn generalizable visual patterns for downstream tasks, followed by fine-tuning the pretrained models to adapt to specific task demands.
Based on this, existing works aim to study: 1) how to collect high-quality person images that benefit model training~\cite{hc_ssp_humanbench, dataset_sapiens}; and 2) how to design a proper pretext task that facilitates the learning of human visual patterns~\cite{hc_ssp_hap,hc_ssp_hcmoco,hc_ssp_solider}. However, all of these works ignore limitations in adapting to real-world applications, which require general visual patterns for downstream tasks, while maintaining a lightweight model architecture with computationally sustainable costs to ensure edge-device compatibility. Such limitations primarily arise from two drawbacks:

\textbf{(1) Specific visual patterns.} In general, human-centric visual perception in real-world scenes often involves numerous instance discrimination tasks and diverse dense prediction tasks, which require general human visual patterns capable of perceiving multiple persons across a broad range of scales. However, existing methods~\cite{hc_ssp_hap, hc_ssp_solider, hc_ssp_hcmoco, ssp_mae_dino, ssp_dino, ssp_mae} pretrain vision models by learning specific visual patterns, i.e., single instance at a single scale. When directly applying these methods to lightweight vision models, a significant misalignment arises between the learned visual patterns and those required for generalization to downstream tasks. As illustrated in Fig.~\ref{fig:domain_change}, when faced with diverse scales of one instance, vision models pretrained using existing methods fail to capture consistent human visual patterns. Additionally, when faced with multiple instances in one image, captured visual patterns appear to focus on the central instance while neglecting the others. This phenomenon suggests that these pretrained models are prone to overfitting specific human visual patterns, i.e., single instance at single scale, thus limiting their generalizability to a wide range of downstream tasks (Fig.~\ref{figure:introduction}(a)).

\textbf{(2) Excessively large model size.} Inspired by the empirical scaling laws~\cite{optim_scaling_laws}, existing works~\cite{hc_ssp_hap, hc_ssp_solider, hc_ssp_hcmoco, hc_ssp_hqnet, hc_ssp_humanbench, hc_ssp_unihcp, dataset_luperson, dataset_sapiens} are based on a flawed assumption: achieving generalizable HVP models requires excessively large model size and extensive datasets of person images. Consequently, HVP models often consist of tens of millions or even billions of parameters, resulting in significant computational costs. However, as depicted in Fig.~\ref{figure:introduction}(a), numerous real-world applications of HVP highly rely on edge devices with limited computational resources, such as pose estimation in the metaverse using VR glasses, person search via surveillance cameras and pedestrian detection in autonomous driving systems. Therefore, existing HVP models with a substantial computational burden are generally incompatible with these edge devices, significantly limiting their practical applicability in real-world scenes. 

To address these limitations, a viable direction is the large-scale pretraining of lightweight model. This motivates us to rethink a significant question that is considerably less studied: \textbf{\textit{is it possible to design a pretraining method that enables a lightweight HVP model to be generalizable for diverse downstream tasks, and how can this be achieved ?}}
In this work, we seek to answer this question by proposing \textbf{S}cale-\textbf{A}ware \textbf{I}mage \textbf{P}retraining (SAIP), a novel self-supervised pretraining framework particularly designed for lightweight HVP models. As illustrated in Fig.~\ref{figure:introduction}(b), deep neural architectures with around 5 millions parameters are utilized as the lightweight vision models. Within the SAIP framework, those lightweight vision models are guided to learn cross-scale consistent patterns under both single-instance and multi-instance settings, aiming to align the learned visual patterns with the requirement of cross-scale consistency in real-world scenes. To this end, three learning tasks are proposed: 1) Cross-scale Matching (CSM) which contrastively learns image-level invariant patterns from multi-scale person-centric images; 2) Cross-scale Reconstruction (CSR) which learns pixel-level consistent patterns from cross-scale masked images; and 3) Cross-scale Search (CSS) which learns to capture diverse patterns from multi-person images across multiple scales. Three tasks complement one another, enabling lightweight models to effectively capture diverse vision patterns such as global discrimination pattern, local structure pattern, and regional discrimination pattern. Consequently, SAIP pretraining facilitates efficient transfer learning for various HVP downstream tasks. 
To summary, the contributions of this paper are three fold.
\begin{itemize}
	\itemsep0em 
	\item \textbf{A new pretraining paradigm.} We propose \textit{SAIP}, the first self-supervised pretraining framework specifically designed to establish lightweight and generalizable models for HVP. Notably, the SAIP effectively utilizes three learning tasks to ensure a coherent alignment across different scales, thereby addressing the requirement of cross-scale consistency in real-world scenarios. 
	\item \textbf{Flexibility and strong generalization.} SAIP is model-agnostic, allowing it to be applied for any model architecture. Based on SAIP, we have successfully established generalizable HVP models with three lightweight architectures (i.e., ViT, ConvNet and the hybrid), and resulting models are comprehensively evaluated across 12 human-centric visual perception datasets. 
	Specifically, SAIP surpasses previous SSP methods across 9 specific human-centric tasks by a large margin. It achieves significant improvements in single-person discrimination tasks by 3\%-13\%, dense predictions by 1\%-11\%, and multi-person visual understanding by 1\%-6\%, exhibiting a remarkable generalization capability.
	\item \textbf{New findings.} Based on extensive experiments, this study reveals that if a proper training objective is explored, a lightweight architecture can serve as a good generalizable model for human-centric visual perception. Additionally, two useful insights for pretraining lightweight vision models are provided: 1) Multi-task self-supervised learning is more beneficial for unlocking lightweight model's potential to learn general human visual patterns; 2) Enhancing the diversity as well as the high fidelity of the dataset, rather than merely increasing the quantity, is beneficial for pretraining.
\end{itemize}

\section{Related Work}
\label{sec:related_works}
\noindent\textbf{Human-centric Visual Perception.}
The intelligent understanding of humans through visual media, such as image and video, has been a long-standing topic in the computer vision community.
Optimally addressing human-centric visual perception would significantly support a wide range of visual applications, such as person ReID in security system, human parsing in metaverse, and person detection in autonomous driving. In general, HVP tasks can be categorized into three groups: 1) Single-person discrimination task (e.g., Person ReID or Pedestrian Attribute Recognition)~\cite{ds_attr_rec_par, ds_reid_transreid, ds_t2ireid_irra}, which focuses on the identity information of a person; 2) Single-person dense predictions task (e.g., pose estimation or body-part segmentation)~\cite{ds_pose_vitpose, ds_hp_parsingrcnn, ds_hp_schp, dataset:LIP}, which particularly focuses on the shape or posture of a person; and 3) multi-person visual understanding task (e.g., multiple human parsing or part-level attribute parsing)~\cite{ds_hp_parsingrcnn, ds_part_parse_kercnn,ds_pdetection_crowddet,dataset:cihp,dataset:fashionpedia}, which is a hybrid task that simultaneously perform person detection and single-person recognition in crowded scene. These tasks are highly correlated as they share a common goal of visually understanding human body structure. Consequently, recent works~\cite{hc_ssp_humanbench, hc_ssp_hcmoco, hc_ssp_unihcp, hc_ssp_hqnet} attempt to unify these tasks by pretraining a large model on extensive publicly available datasets with multi-task labels. However, the scarcity of high-quality labels significantly limits the scalability of the pretrained models.

In contrast to existing works, proposed SAIP is built on large-scale self-supervised pretraining framework, which can leverage massive public unlabeled data. Furthermore, this work does not seek to attain the performance upper bound of large models; instead, it explores the potential of lightweight models, which are essential for broadly supporting human-centric visual perception.

\noindent\textbf{Self-supervised Visual Pretraining.} 
Learning meaningful patterns from massive unlabeled data has been shown to be a promising approach for developing visual foundation models. As presented in Fig~\ref{figure:introduction}.(a), mainstream studies in general can be categorized into two paradigms: contrastive learning (CL) \cite{ssp_dino,ssp_moco,hc_ssp_hcmoco,hc_ssp_solider,ssp_oadp} and masked image modeling (MIM) \cite{ssp_mae,ssp_eva,ssp_mae-lite,ssp_tinymim,hc_ssp_hap}. The CL models image similarity and dissimilarity between two or more views of the same object, aiming to capture long-range global patterns such as shape. In contrast, the MIM focuses on reconstruction of masked images, preferring high-frequency signals that are texture-oriented. Recent studies~\cite{ssp_mae_dino} reveal that CL and MIM can complement each other and their straightforward combination can enhance model pretraining. 

In contrast to these methods that primarily focus on specific characteristic of an object for pretraining, we take a distinct approach: our method introduces scale-aware image pretraining, which particularly learns cross-scale consistent representations from both image- and pixel-level perspective. Our core finding is that a proper learning strategy such as SAIP can unlock the potential of lightweight models (around 5 million parameters), and thus avoid heavy burden of optimization of tens millions or even billions parameters.

\section{Method}
\label{sec:method}
\begin{figure*}
	\centering
	\includegraphics[width=0.9\linewidth]{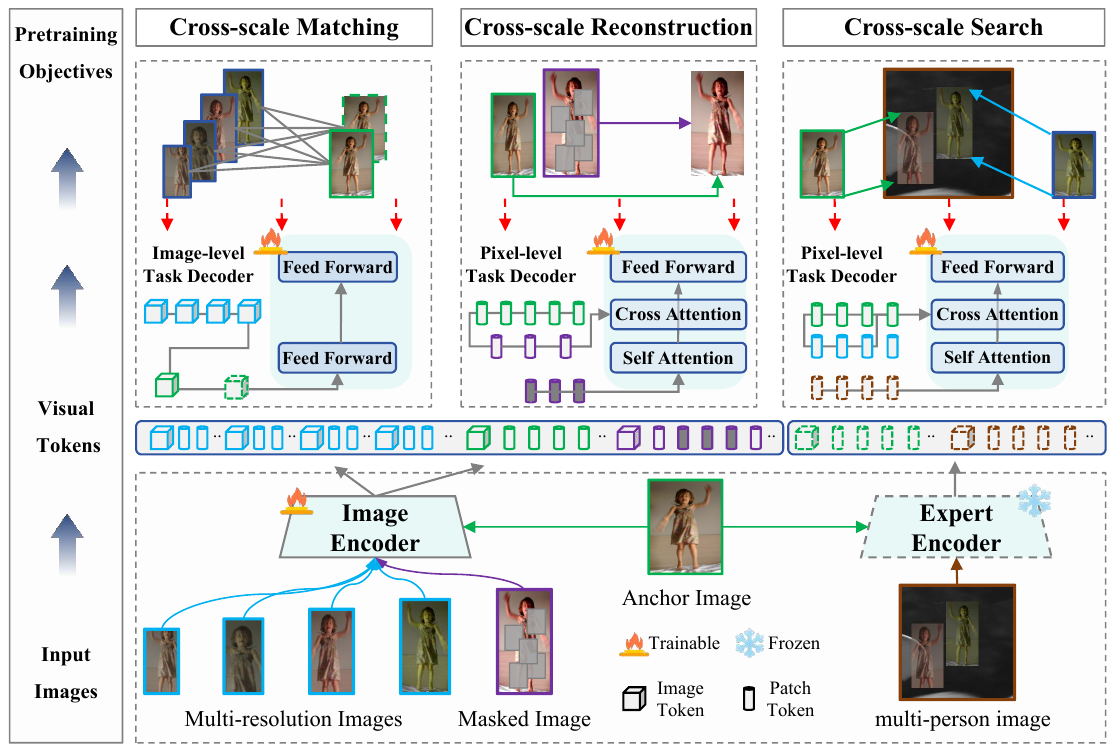}
	\caption{The overview of \textbf{S}cale-\textbf{A}ware \textbf{I}mage \textbf{P}retraining (SAIP), a self-supervised pretraining framework to learn human visual patterns from large scale person images. The framework employs a hybrid architecture comprising an image encoder that tokenize input images into visual tokens, an expert encoder that provides ground truth labels, and three distinct task decoders, each corresponding to one of the three specific learning objectives.}
	\vspace*{-0.1in}
	\label{figure:method}
\end{figure*}

We propose SAIP, as illustrated in Fig.~\ref{figure:method}, a self-supervised pretraining framework designed to learn human-centric representations from large scale person images. This section first introduces model architecture, and subsequently delineates learning objectives.

\begin{table*}[ht]
	\centering
	\resizebox{0.99\linewidth}{!}{
		\renewcommand{\arraystretch}{1.3}
		\begin{tabular}{l|l|c|c|cc|cc|cc}
			\hline
			\multicolumn{1}{c|}{\multirow{2}{*}{Method}} & Vision      & Model & Pre-train                 & \multicolumn{2}{c|}{I2I ReID} & \multicolumn{2}{c|}{T2I ReID} & \multicolumn{2}{c}{Attribute Recognition} \\
			\multicolumn{1}{c|}{}                        & backbone    & size  & dataset                   & Market $\uparrow$         & MSMT17 $\uparrow$        & CUHK   $\uparrow$       & ICFG   $\uparrow$       & PA100K   $\uparrow$            & PETAzs $\uparrow$             \\ 
			\hline		
			\multicolumn{1}{c|}{\multirow{5}{*}{\begin{tabular}[c]{@{}l@{}}Supervised\\ Pretraining\end{tabular} }}                                    & StarNet\_S3~\cite{arch_cnn_starnet} & 5.8M   & \multicolumn{1}{c|}{IN1k} & \textbf{90.5}          & 65.8        & 55.4       & 44.8        & 79.1               & \textbf{71.9}              \\
			& CSPNeXt\_S~\cite{arch_cnn_cspnext}   & 4.4M   & \multicolumn{1}{c|}{AIC+COCO} & 89.8          & 63.6         & 47.5        & 33.6        & 78.9               & 67.3             \\ 
			& ViT\_Tiny~\cite{arch_vit_vit}   & 5.5M   & \multicolumn{1}{c|}{IN1K} & 89.3          & 65.9       & 51.1        & 45.9        & 74.5               & 68.0              \\ 
			& EdgeNeXt~\cite{arch_hybrid_edgenext}   & 5.6M   & \multicolumn{1}{c|}{IN1K} & 89.5         & 67.5         & \textbf{58.0}        & \textbf{47.4}        & \textbf{80.6}               & 70.8              \\ 
			& TinyViT\_5M~\cite{arch_hybrid_tinyvit}   & 5.4M   & \multicolumn{1}{c|}{IN1K} & 89.1          & \textbf{68.9}         & 57.8        & 40.9        & 76.5               & 69.5              \\ 
			\hline
			\multicolumn{10}{l}{\begin{tabular}[l]{@{}l@{}}Self-supervised Pretraining\end{tabular}} \\
			\hline
			DINO~\cite{ssp_dino}                             & ViT\_Tiny~\cite{arch_vit_vit} & 5.5M   & \multicolumn{1}{c|}{LUP1M} & 90.5          & 65.8         & 55.3        & 40.2        & 77.4               & 69.3              \\
			MAE~\cite{ssp_mae}	& ViT\_Tiny~\cite{arch_vit_vit}   & 5.5M    & \multicolumn{1}{c|}{LUP1M} & 79.7          & 39.9         & 36.6       & 19.1       & 68.3               & 61.1             \\ 
			MAE+DINO~\cite{ssp_mae_dino}	& ViT\_Tiny~\cite{arch_vit_vit}   & 5.5M    & \multicolumn{1}{c|}{LUP1M} & 89.2          & 61.6         & 52.7        & 37.7        & 72.9               & 66.5              \\ 
			HAP~\cite{hc_ssp_hap}	& ViT\_Tiny~\cite{arch_vit_vit}   & 5.5M    & \multicolumn{1}{c|}{LUP1M} & 81.6          & 42.4        & 40.2        & 20.4        & 66.3               & 64.1              \\ 
			SOLIDER~\cite{hc_ssp_solider}	& ViT\_Tiny~\cite{arch_vit_vit}   & 5.5M    & \multicolumn{1}{c|}{LUP1M} & 91.6          & 69.2         & 55.5        & 40.7       & 78.6               & 69.4              \\ 
			\textbf{SAIP (Ours)}	& ViT\_Tiny~\cite{arch_vit_vit}   & 5.5M    & \multicolumn{1}{c|}{LUP1M} & 93.6          & 75.6         & 59.2        & 46.1        & 80.7               & 71.4              \\ 
			\textbf{SAIP* (Ours)}	& ViT\_Tiny~\cite{arch_vit_vit}   & 5.5M    & \multicolumn{1}{c|}{LUP1M} & \textbf{95.2}          & \textbf{83.1}         & \textbf{65.9}        & \textbf{56.9}        & \textbf{82.7}              & \textbf{72.4}             \\ 
			\hline
			DINO~\cite{ssp_dino}                             & StarNet\_S3~\cite{arch_cnn_starnet} & 5.8M    & \multicolumn{1}{c|}{LUP1M} & 89.1          & 60.6         & 48.5        & 36.1        & 76.2               & 68.8              \\
			SOLIDER~\cite{hc_ssp_solider}	& StarNet\_S3~\cite{arch_cnn_starnet} & 5.8M    & \multicolumn{1}{c|}{LUP1M} & 89.9          & 61.9         & 49.9        & 38.4        & 77.4               & 69.3              \\ 
			\textbf{SAIP (Ours)}	& StarNet\_S3~\cite{arch_cnn_starnet} & 5.8M    & \multicolumn{1}{c|}{LUP1M} & 92.0          & 66.1         & 52.9        & 41.3        & 78.1               & 70.9             \\ 
			\textbf{SAIP* (Ours)}	& StarNet\_S3~\cite{arch_cnn_starnet} & 5.8M    & \multicolumn{1}{c|}{LUP1M} & \textbf{95.3}          & \textbf{83.4}         & \textbf{65.4}        & \textbf{56.7}       & \textbf{83.6}              & \textbf{73.2}             \\ 
			\hline
			DINO~\cite{ssp_dino}                             & TinyViT\_5M~\cite{arch_hybrid_tinyvit} & 5.4M     & \multicolumn{1}{c|}{LUP1M} & 89.4          & 57.9        & 46.5        & 35.0        & 78.6               & 70.1             \\
			SOLIDER~\cite{hc_ssp_solider}	& TinyViT\_5M~\cite{arch_hybrid_tinyvit} & 5.4M     & \multicolumn{1}{c|}{LUP1M} & 89.8          & 59.9         & 47.1        & 34.9        & 79.0               & 70.3             \\ 
			\textbf{SAIP (Ours)}	& TinyViT\_5M~\cite{arch_hybrid_tinyvit} & 5.4M    & \multicolumn{1}{c|}{LUP1M} & 94.1         & 73.5         & 59.6       & 47.7        & 82.1               & 72.8             \\ 
			\textbf{SAIP* (Ours)}	& TinyViT\_5M~\cite{arch_hybrid_tinyvit} & 5.4M     & \multicolumn{1}{c|}{LUP1M} & \textbf{96.0}          & \textbf{84.6}         & \textbf{66.2}        & \textbf{53.2}        & \textbf{83.5}               & \textbf{74.8}              \\ 
			\hline
		\end{tabular}
	}
	\caption{Quantitative comparison with state-of-the-art pretraining methods on 6 single-person discrimination datasets. * denotes the expert encoder is a pretrained model such as PATH~\cite{hc_ssp_humanbench}. $\uparrow$ means the larger value the better performance.}
	\vspace*{-1.em}
	\label{table:sota_id_tasks}
\end{table*}
\subsection{Model Architecture}
In order to pretrain a lightweight model with strong capability of understanding persons from cross-scale perspectives, we propose a hybrid architecture that explores multi-task learning. Inspired by previous works~\cite{ssp_dino,ssp_moco}, the pretraining paradigm is founded on the idea of knowledge distillation, which consists of an image encoder, an expert encoder, and three task decoders. In particular, the expert encoder is solely used to construct learning targets for stabilizing model training. After training, the expert encoder and task decoders are discarded, while only the image encoder is retained for downstream tasks.

\noindent\textbf{(1) Image encoder,} a visual feature extractor designed to tokenize an image into numerous visual tokens. The image encoder can be built on various architectures, such as vision transformer (ViT), convolution neural network (CNN) or even a hybrid of ViT and CNN. In this work, we adopt plain vision transformer~\cite{arch_vit_vit}, StarNet~\cite{arch_cnn_starnet} and TinyViT~\cite{arch_hybrid_tinyvit} as the instantiations respectively for ViT-, CNN- and hybrid-based image encoder. Formally, given an image $I \in \mathbb{R}^{3\times H\times W}$, the image encoder outputs an image token $F \in \mathbb{R}^{D}$ representing entire image and $L$ patch tokens $F^{\textquotesingle} \in \mathbb{R}^{L \times D}$ as well.

\noindent\textbf{(2) Expert encoder,} a visual feature extractor specifically for providing learning targets only. The expert encoder can either be a moving-averaged encoder, whose parameters are updated with an exponential moving average of image encoder, or an existing model that has been pretrained on large scale data.

\noindent\textbf{(3) Task decoders,} which utilize image-level decoder or pixel-level decoder to perform three proxy tasks. In particular, the image-level decoder is a stack of feed-forward layers that project the image token $F$ to instance representation $\hat{F} \in \mathbb{R}^{d}$ for the purpose of cross-scale matching. The pixel-level decoder, denoted as $\phi([query], [key], [value])$, comprises a stack of interactive computation blocks, each of which sequentially includes self-attention layer, cross-attention layer and feed-forward layer. Two types of image pair is considered for the pixel-level decoder: 1) masked image and non-masked image; and 2) single-instance image and multi-instances image. The pixel-level decoder take the former pair as input for cross-scale reconstruction task, and the latter pair for cross-scale search.

\subsection{Pretraining Objectives}
We optimize lightweight models using three learning objectives. To facilitate the pretraining process, we prepare three types of inputs: an anchor image $I_0$ of a person, $N$ images with different resolutions for the person $\{I_t\}_{t=1}^{N}$, and a multi-person image $I_{N+1}$. Specifically, the multi-resolution images are obtained by randomly resizing the anchor image to different scales ranging from 0.75 to 1.5, while the multi-person image is synthesized by applying a simple copy-paste technique~\cite{data_aug_copypaste} to multi-resolution images.
Based on these different inputs, task decoders are activated to compute the three losses, as delineated below. 

\noindent\textbf{(1) Cross-scale Matching (CSM)} designed to constrastively learn multi-scale invariant representations. Given the anchor image and multi-resolution images, the image-level decoder is activated to obtain $N+1$ instance representations, and the consistency between them is computed using cross-entropy loss as formulated in Equ~\ref{equ.loss_crm}:
\begin{equation}
	\begin{array}{lll}
		\ell_{csm} & \hspace{-0.5em} = & \hspace{-0.5em} \frac{1}{N}\sum\limits_{t}^{N} -\hat{F_{0}} log(\hat{F_{t}})\\ 
	\end{array}
	\label{equ.loss_crm} 
\end{equation}
where $\hat{F_{0}}$ denotes the instance representation of the anchor image extracted by image encoder with image-level decoder, while $\hat{\{F_t\}}^{N}_{t=1}$ denotes the counterpart of $N$ multi-resolution images which are obtained by the expert encoder and image-level decoder.

\noindent\textbf{(2) Cross-scale Reconstruction (CSR)} designed to learn common structure of a person instance at different scales. To achieve this, we utilize the pixel-level decoder with cross-attention to reconstruct a masked image conditioned on anchor image. Specifically, we randomly select one of multi-resolution images and divide it into non-overlapping patches. A subset of these patches (e.g., mask ratio is 75\% in general) is randomly masked, leaving the rest visible. Regarding the masked image as the query, the pixel-level decoder is activated to project patch tokens of anchor image to RGB space for the reconstruction of the masked image. The similarity between the predicted image and selected image is measured by mean square errors as formulated in Equ~\ref{equ.loss_crr}:
\begin{equation}
	\begin{array}{lll}
		\ell_{csr} &\hspace{-0.5em} = & \hspace{-0.2em} \left\| \phi(\widetilde{F}_{t}^{\textquotesingle}, F_{0}^{\textquotesingle}, F_{0}^{\textquotesingle}) - I_{t} \right\|^2_2  \\ 
	\end{array}
	\label{equ.loss_crr} 
\end{equation}
where $F_{0}^{\textquotesingle}$ denotes the patch tokens of the anchor image which are extracted by image encoder, while $\widetilde{F}_{t}^{\textquotesingle}$ denotes patch tokens of the masked image which are extracted by expert encoder.

\noindent\textbf{(3) Cross-scale Search (CSS),} which aims to learn fine-grained alignment between single instance and multi-instances by identifying a specific instance within a multi-person image. To achieve this, we introduce a coarse instance segmentation task, which requires pixel-decoder to segment area corresponding to the given instance. Regarding the multi-resolutions images as the query, the CSS learning objective is formulated in Equ~\ref{equ.loss_crs}:
\begin{equation}
	\begin{array}{lll}
		\ell_{css} & \hspace{-0.5em} = &\hspace{-0.2em} \sum_{t=1}^{N}\left\| \tau(\phi(F_{t}^{\textquotesingle}, F_{N+1}^{\textquotesingle}, F_{N+1}^{\textquotesingle})) - M_{t} \right\|^2_2  \\ 
	\end{array}
	\label{equ.loss_crs} 
\end{equation}
where $F_{t}^{\textquotesingle}$ denotes patch tokens of $t$-th multi-resolutions image which are extracted by image encoder, while $F_{N+1}^{\textquotesingle}$ denotes the patch tokens of the multi-person image which are extracted by expert encoder. $\tau(\cdot)$ is the segmentation function that predicts binary mask for each instance, and $M_{t}$ is the ground-truth binary mask indicating the area of $t$-th instance.

The overall learning objective $\mathcal{L}$ can thus be expressed as follows in Equ~\ref{equ.saip_loss}:
\begin{equation}
	\begin{array}{lll}
		& \mathop{min}\limits_{\theta} \mathbb{E}[\mathcal{L}(I_0, I_t, I_{N+1})],  & \\ 
		& s.t., \,\,\, \mathcal{L}=\ell_{csm}+\ell_{csr}+\ell_{css} &
	\end{array}
	\label{equ.saip_loss} 
\end{equation}
where $\theta$ is the learnable parameters of a lightweight model.

\begin{table*}[ht]
	\centering
	\renewcommand\arraystretch{1.3}
	\resizebox{0.95\linewidth}{!}{
		\begin{tabular}{l|l|c|c|cc|cc|cc}
			\hline
			\multicolumn{1}{c|}{\multirow{2}{*}{Method}} & Vision      & Model & Pre-train                 & \multicolumn{2}{c|}{Pose Estimation} & \multicolumn{2}{c|}{Landmark Detection} & \multicolumn{2}{c}{Part Segmentation} \\
			\multicolumn{1}{c|}{}                        & backbone    & size  & dataset                   & $AP \uparrow$          & $AR \uparrow$         &   $AP \uparrow$          & $AR \uparrow$          & $mIoU \uparrow$               & $mAcc \uparrow$              \\ 
			\hline
			\multicolumn{1}{c|}{\multirow{5}{*}{\begin{tabular}[c]{@{}l@{}}Supervised\\ Pretraining\end{tabular} }}                                    & StarNet\_S3~\cite{arch_cnn_starnet} & 5.8M   & \multicolumn{1}{c|}{IN1k} & \textbf{71.9}          &  \textbf{75.1}        &    50.5     &   62.4     & 52.4               &      61.9         \\
			& CSPNeXt\_S~\cite{arch_cnn_cspnext}   & 4.4M   & \multicolumn{1}{c|}{AIC+COCO} & 70.0          & 73.3         &   50.3     &   62.3      & 54.6               &     63.6          \\ 
			& ViT\_Tiny~\cite{arch_vit_vit}    & 5.5M   & \multicolumn{1}{c|}{IN1K} & 69.3          & 72.6         &  45.0      &     58.1    & 48.9              &      63.5        \\ 
			& EdgeNeXt~\cite{arch_hybrid_edgenext}   & 5.6M   & \multicolumn{1}{c|}{IN1K} &     71.6      &   74.8       &    50.5     &    62.2     &     54.9          &    65.5           \\ 
			& TinyViT\_5M~\cite{arch_hybrid_tinyvit}   & 5.4M   & \multicolumn{1}{c|}{IN1K} & 69.2         & 72.7        &  \textbf{53.7}     &     \textbf{65.0}   & \textbf{56.2}               &       \textbf{67.7}        \\ 
			\hline
			\multicolumn{10}{l}{\begin{tabular}[l]{@{}l@{}}Self-supervised Pretraining\end{tabular}} \\
			\hline
			DINO~\cite{ssp_dino}                            & ViT\_Tiny~\cite{arch_vit_vit}  & 5.5M   & \multicolumn{1}{c|}{LUP1M} & 69.3          &    72.6     &    43.9     &    57.1    & 48.7               &     59.3        \\
			MAE~\cite{ssp_mae}	& ViT\_Tiny~\cite{arch_vit_vit}   & 5.5M   & \multicolumn{1}{c|}{LUP1M} & 67.0          &  70.6        &    40.1     &   52.7      & 43.5               &        54.0      \\ 
			MAE+DINO~\cite{ssp_mae_dino}	& ViT\_Tiny~\cite{arch_vit_vit}    & 5.5M   & \multicolumn{1}{c|}{LUP1M} & 69.9          &  73.2       &     45.1    &   58.2     & 49.8               &   60.3            \\ 
			HAP~\cite{hc_ssp_hap}	& ViT\_Tiny~\cite{arch_vit_vit}    & 5.5M   & \multicolumn{1}{c|}{LUP1M} & 68.8          &    72.3      &    42.6     &    55.4     & 44.4               &     54.5        \\ 
			SOLIDER~\cite{hc_ssp_solider}	& ViT\_Tiny~\cite{arch_vit_vit}    & 5.5M   & \multicolumn{1}{c|}{LUP1M} & 69.3          &    72.6      &     44.2    &   57.2      & 48.9              &   59.2           \\ 
			\textbf{SAIP (Ours)}	& ViT\_Tiny~\cite{arch_vit_vit}    & 5.5M   & \multicolumn{1}{c|}{LUP1M} & 70.1         &  73.3       &    45.7     &     58.7    & 52.3               &    63.3         \\ 
			\textbf{SAIP* (Ours)}	& ViT\_Tiny~\cite{arch_vit_vit}    & 5.5M   & \multicolumn{1}{c|}{LUP1M} & \textbf{70.9}          &   \textbf{74.1}     &   \textbf{46.3}      &    \textbf{59.6}    & \textbf{55.2}              &  \textbf{67.6}             \\ 
			\hline
			DINO~\cite{ssp_dino}                            & StarNet\_S3~\cite{arch_cnn_starnet}  & 5.8M   & \multicolumn{1}{c|}{LUP1M} & 71.1          &   74.3       &     49.5    &     60.8    & 51.4               &       60.1       \\
			SOLIDER~\cite{hc_ssp_solider}	& StarNet\_S3~\cite{arch_cnn_starnet}  & 5.8M   & \multicolumn{1}{c|}{LUP1M} & 71.5          &    74.5     &  49.9      &   61.3      & 52.1               &      60.8       \\ 
			\textbf{SAIP (Ours)}	& StarNet\_S3~\cite{arch_cnn_starnet}  & 5.8M   & \multicolumn{1}{c|}{LUP1M} & \textbf{72.7}          &   \textbf{75.7}       &     50.0    &    61.6     & 52.7              &     61.7       \\ 
			\textbf{SAIP* (Ours)}	& StarNet\_S3~\cite{arch_cnn_starnet}  & 5.8M   & \multicolumn{1}{c|}{LUP1M} & 72.3          &   75.5       &    \textbf{52.7}    &  \textbf{64.2}     & \textbf{54.3}               &    \textbf{63.6}           \\ 
			\hline
			DINO~\cite{ssp_dino}                            & TinyViT\_5M~\cite{arch_hybrid_tinyvit} & 5.4M   & \multicolumn{1}{c|}{LUP1M} & 65.4          &  68.8      &    51.6    &   62.8     & 55.9               &    66.0           \\
			SOLIDER~\cite{hc_ssp_solider}	& TinyViT\_5M~\cite{arch_hybrid_tinyvit} & 5.4M   & \multicolumn{1}{c|}{LUP1M} & 65.7          &  69.0       &   51.5     &   62.7     & 56.2               &   66.7         \\ 
			\textbf{SAIP (Ours)}	& TinyViT\_5M~\cite{arch_hybrid_tinyvit} & 5.4M   & \multicolumn{1}{c|}{LUP1M} & 72.1          &  75.0       &   53.9      &   65.2      & \textbf{58.3}               &    \textbf{68.8}          \\ 
			\textbf{SAIP* (Ours)}	& TinyViT\_5M~\cite{arch_hybrid_tinyvit} & 5.4M   & \multicolumn{1}{c|}{LUP1M} & \textbf{73.3}          &   \textbf{76.2}       &   \textbf{54.2}     &  \textbf{65.5}       & 57.0               &    67.9        \\ 
			\hline
		\end{tabular}
	}
	\caption{Quantitative comparison with state-of-the-art pretraining methods on 3 single-person dense predictions datasets.}
	\vspace*{-1.em}
	\label{table:sota_dp_tasks}
\end{table*}
\section{Experiments}
\label{sec:experiments}
In this section, we first present an overview of the experimental settings, followed by the implementation details. Subsequently, we analyze the merit of SAIP through a comprehensive comparison with existing methods on 9 human-centric tasks, which include image-to-image person ReID (I2I ReID), text-to-image person ReID (T2I ReID), person attribute recognition (PAR), human pose estimation (HPE), person landmarks detection (PLD), body part segmentation (BPS), pedestrian detection (PD), multi-person human parsing (MHP), and part-level attribute parsing (PAP). Finally, we conduct comprehensive ablation study to explore major properties of proposed SAIP.

\subsection{Experimental Settings}
\noindent \textbf{Datasets.} Unless otherwise specified, all self-supervised models in this paper are pretrained on LUP1M, which comprises 1 million person images randomly selected from LUPerson dataset~\cite{dataset_luperson}. In terms of evaluation on downstream tasks, the pretrained models are fully tested on commonly-used datasets. Specifically, we adopt Market1501~\cite{dataset:market_reid} and MSMT17~\cite{dataset:msmt_reid} for I2I ReID, CUHK-PEDES~\cite{dataset:t2i_reid_cuhk} and ICFG-PEDES~\cite{dataset:t2i_reid_icfg} for T2I ReID, PA100K~\cite{dataset:pa100k} and PETA~\cite{dataset:peta} for PAR, COCO-Keypoints~\cite{dataset:coco} for HPE, Whole-body COCO~\cite{dataset:wholebody-coco} for PLD, LIP~\cite{dataset:LIP} for BPS, CrowdHuman~\cite{dataset:crowdhuman} for PD, CIHP~\cite{dataset:cihp} for MHP and Fashionpedia~\cite{dataset:fashionpedia} for PAP.

\noindent \textbf{Evaluation Metric.} Following previous works, we adopt $Rank1$ as the evaluation metric for I2I ReID and T2I ReID, mean accuracy ($mA$) for PAR. As for the dense predictions tasks, average precision ($AP$) and recall ($AR$) are used for HPE and PLD, while mean intersection of union ($mIoU$) and mean pixel accuracy ($mAcc$) are applied for BPS. In terms of multi-person understanding tasks, we adopt missing rate ($MR$) and $AP$ for PD, $mIoU$ and $AP$ for MHP, $AP^{box}_{IoU+F_{1}}$ and $AP^{segm}_{IoU+F_{1}}$ for PAP, respectively.

\subsection{Implementation Details}
\noindent \textbf{Pretraining.} All lightweight models are pretrained or finetuned using 8 A6000 48G GPUs. We use the AdamW~\cite{optim_adamw} as the optimizer and set the effective batch size is 2048 (i.e., 256 per GPU). Each model is pretrained from scratch for 300 epochs in default, with a initial learning rate of 2.5e-4, which is declined by Cosine Annealing scheduler~\cite{optim_cosine_anneal}. Other settings are identical to that of DINO~\cite{ssp_dino}.

\noindent\textbf{Finetuning.} For evaluation of pretrained methods, we adopt representative solutions in downstream tasks as the baselines, in which we replace their backbones with different lightweight models that are pretrained via different supervised/self-supervised approaches. 
For example, we adopt TransReID~\cite{ds_reid_transreid} as the basic solution for I2I ReID, IRRA~\cite{ds_t2ireid_irra} for T2I ReID and PA~\cite{dataset:pa100k} for PAR.

\begin{table*}[t]
	\centering
	\renewcommand\arraystretch{1.3}
	\resizebox{0.99\linewidth}{!}{
		\begin{tabular}{l|l|c|c|cc|cc|cc}
			\hline
			\multicolumn{1}{c|}{\multirow{2}{*}{Method}} & Vision      & Model & Pre-train                 & \multicolumn{2}{c|}{Pedestrian Detection} & \multicolumn{2}{c|}{Multiple Human Parsing} & \multicolumn{2}{c}{Part-level Attribute Parsing} \\
			\multicolumn{1}{c|}{}                        & backbone    & size  & dataset                   & $AP \uparrow$          & $MR \downarrow$            & $mIoU \uparrow$   & $AP_p \uparrow$       & $AP^{box}_{IoU+F_{1}} \uparrow$               & $AP^{segm}_{IoU+F_{1}} \uparrow$              \\ 
			\hline
			\multicolumn{1}{c|}{\multirow{5}{*}{\begin{tabular}[c]{@{}l@{}}Supervised\\ Pretraining\end{tabular} }}                                    & StarNet\_S3~\cite{arch_cnn_starnet} & 5.8M   & \multicolumn{1}{c|}{IN1k} & 89.2          &  45.3           &  50.9  &    49.4     &     36.9           &   33.8            \\
			& ResNet18~\cite{arch_cnn_resnet}   & 11M   & \multicolumn{1}{c|}{IN1K} &    86.9      &  48.7        &   51.3     &  49.6       & 35.2              &    32.8           \\ 
			& ViT\_Tiny~\cite{arch_vit_vit}    & 5.5M   & \multicolumn{1}{c|}{IN1K} &   86.9       &     50.0    &    47.5    &  46.1       &     37.5           &     34.8         \\ 
			& EdgeNeXt~\cite{arch_hybrid_edgenext}   & 5.6M   & \multicolumn{1}{c|}{IN1K} &     \textbf{89.6}      &   \textbf{44.3}       &     \textbf{53.0}    &    \textbf{51.3}     &     \textbf{40.1}          &   \textbf{37.1}            \\ 
			& TinyViT\_5M~\cite{arch_hybrid_tinyvit}    & 5.4M   & \multicolumn{1}{c|}{IN1K} &   88.9       &    44.6      &  51.7      &    50.4    &   38.5           &     35.3          \\ 
			\hline
			\multicolumn{10}{l}{\begin{tabular}[l]{@{}l@{}}Self-supervised Pretraining\end{tabular}} \\
			\hline
			DINO~\cite{ssp_dino}                             & ViT\_Tiny~\cite{arch_vit_vit}  & 5.5M   & \multicolumn{1}{c|}{LUP1M} &   86.1      &  51.6       &   46.9      &    45.7    &      35.4         &     32.9        \\
			MAE~\cite{ssp_mae}	& ViT\_Tiny~\cite{arch_vit_vit}    & 5.5M   & \multicolumn{1}{c|}{LUP1M} &    83.7      &    56.5      &  45.8       &  44.4       &      32.0        &      30.3        \\ 
			MAE+DINO~\cite{ssp_mae_dino}	& ViT\_Tiny~\cite{arch_vit_vit}    & 5.5M   & \multicolumn{1}{c|}{LUP1M} &   86.4        &  50.2      &  47.4       &  46.2      & 37.7             &  35.2             \\ 
			HAP~\cite{hc_ssp_hap}	& ViT\_Tiny~\cite{arch_vit_vit}    & 5.5M   & \multicolumn{1}{c|}{LUP1M} &      83.3    &    57.5     &    44.1     &  42.8       & 33.0               &    30.4         \\ 
			SOLIDER~\cite{hc_ssp_solider}	& ViT\_Tiny~\cite{arch_vit_vit}    & 5.5M   & \multicolumn{1}{c|}{LUP1M} &   85.7      &    51.9      &    46.8     &   45.7      & 36.7              &    34.2          \\ 
			\textbf{SAIP (Ours)}	& ViT\_Tiny~\cite{arch_vit_vit}    & 5.5M   & \multicolumn{1}{c|}{LUP1M} &   87.1     &   49.6       &     48.2    &  46.9       &  \textbf{38.0}             &   \textbf{35.5}           \\ 
			\textbf{SAIP* (Ours)}	& ViT\_Tiny~\cite{arch_vit_vit}    & 5.5M    & \multicolumn{1}{c|}{LUP1M} &    \textbf{87.2}       &   \textbf{49.4}      &   \textbf{49.2}    &  \textbf{47.8}       &  \textbf{38.0}             &   35.4          \\
			\hline
			DINO~\cite{ssp_dino}                             & StarNet\_S3~\cite{arch_cnn_starnet} & 5.8M    & \multicolumn{1}{c|}{LUP1M} &    86.6       &  48.7        &    50.2     &  49.1       &     32.7          &      29.9        \\
			SOLIDER~\cite{hc_ssp_solider}	& StarNet\_S3~\cite{arch_cnn_starnet} & 5.8M    & \multicolumn{1}{c|}{LUP1M} &      86.1   &     52.1    &  50.0      &    49.1     &    34.3             &     31.5        \\ 
			\textbf{SAIP (Ours)}	& StarNet\_S3~\cite{arch_cnn_starnet} & 5.8M    & \multicolumn{1}{c|}{LUP1M} &    88.1      &  47.4        &   50.3      &   48.6      &       33.9        &   31.3         \\ 
			\textbf{SAIP* (Ours)}	& StarNet\_S3~\cite{arch_cnn_starnet} & 5.8M   & \multicolumn{1}{c|}{LUP1M} &   \textbf{88.8}        &  \textbf{45.5}       &    \textbf{52.9}    &  \textbf{51.2}     &     \textbf{37.9}          &   \textbf{35.0}            \\ 
			\hline
			DINO~\cite{ssp_dino}                             & TinyViT\_5M~\cite{arch_hybrid_tinyvit}  & 5.4M   & \multicolumn{1}{c|}{LUP1M} &    88.9      &    43.6     &   49.8     &  48.9      & 36.1              &   33.4            \\
			SOLIDER~\cite{hc_ssp_solider}	& TinyViT\_5M~\cite{arch_hybrid_tinyvit}  & 5.4M   & \multicolumn{1}{c|}{LUP1M} &     89.5     &  43.3       &  49.9      &   48.9     & 35.9             &    33.2        \\ 
			\textbf{SAIP (Ours)}	& TinyViT\_5M~\cite{arch_hybrid_tinyvit}  & 5.4M   & \multicolumn{1}{c|}{LUP1M} &   90.5       &  \textbf{40.4}        &   51.8      &   50.7      & 39.2              &     35.8         \\ 
			\textbf{SAIP* (Ours)}	& TinyViT\_5M~\cite{arch_hybrid_tinyvit}  & 5.4M   & \multicolumn{1}{c|}{LUP1M} &    \textbf{90.6}       &    41.3     &   \textbf{53.4}     &     \textbf{52.2}    &     \textbf{41.3}          &    \textbf{38.1}        \\ 
			\hline
		\end{tabular}
	}
	\caption{Quantitative comparison with state-of-the-art pretraining methods on 3 multi-person visual understanding tasks. $\downarrow$ means the smaller value the better performance.}
	\vspace*{-1.em}
	\label{table:sota_mpvu_tasks}
\end{table*}

\begin{table}[ht]
	\centering
	\renewcommand\arraystretch{1.1}
	\resizebox{0.99\linewidth}{!}{
		\begin{tabular}{ccc|c|cccc}
			\hline
			\multicolumn{3}{c|}{Pretraining} &\begin{tabular}[c]{@{}c@{}}Pretrain \\ dataset \end{tabular} & \begin{tabular}[c]{@{}c@{}}ReID \\ ($Rank1$) \end{tabular} & \begin{tabular}[c]{@{}c@{}}PAR \\ ($mA$) \end{tabular}  & \begin{tabular}[c]{@{}c@{}}HP \\ ($mIoU$) \end{tabular} \\ 
			\hline
			\multicolumn{3}{c|}{None } & -	& 77.1  &  65.8  & 40.6 \\
			\multicolumn{3}{c|}{Supervised}	& IN1K  & 89.3  &  74.5  & 48.9 \\
			\hline
			CSM & CSR & CSS & & &  \\ 
			\hline
			\checkmark	&     &     & LUP1M & 92.3 &  78.6     &   49.6        \\ 
			&   \checkmark  &     &  LUP1M &  87.1 & 72.4     &   45.2        \\ 
			&     &   \checkmark  &   LUP1M&  89.1 & 75.2     &   47.8        \\ 
			\checkmark	& \checkmark    &  & LUP1M  &  93.3 &   80.4     &   51.1        \\ 
			\checkmark	&     &  \checkmark   & LUP1M & 92.2 &   79.1     &   50.7        \\ 
			\checkmark	&  \checkmark   & \checkmark    & LUP1M &  93.6 &  80.7     &   52.3        \\ 
			\hline
		\end{tabular}
	}
	\caption{Investigation of the effect of learning tasks.}
	\vspace*{-0.5em}
	\label{table:abla_saip_component}
\end{table}
\begin{table}[t]
	\centering
	\renewcommand\arraystretch{1.1}
	\resizebox{0.99\linewidth}{!}{
		\begin{tabular}{l|c|ccc}
			\hline
			\begin{tabular}[c]{@{}l@{}}Pretrain  dataset \end{tabular} & \#images & \begin{tabular}[c]{@{}c@{}}ReID \\ ($Rank1$) \end{tabular} & \begin{tabular}[c]{@{}c@{}}PAR \\ ($mA$) \end{tabular}  & \begin{tabular}[c]{@{}c@{}}HP \\ ($mIoU$) \end{tabular} \\
			\hline
			IN1K                                                       & 1284228  & 90.2 & 77.8   &  50.1  \\
			HP1M                                                       & 1218136  & 93.9 & 80.9   &  53.0   \\
			\hline
			LUP1M                                                      & 1284228  & 93.6 & 80.7   &  52.3  \\
			LUP2M                                                      & 2568456  & 93.7 & 80.4   &  52.6  \\
			LUP4M                                                      & 4180243  & 93.9 & 80.9   &  52.4  \\
			\hline
			LUP1M+IN1K                                                 & 2568456  & 93.8 & 81.4   &  52.4 \\
			LUP1M+HP1M                                                 & 2502364  & 94.0 & 81.7   &  52.9 \\
			LUP1M+HP1M+IN1K                                            & 3783531  & 94.4 & 82.3   &  53.6 \\
			\hline
		\end{tabular}
	}
	\caption{Investigating the impact of the data from three aspects: quantity, diversity and quality. Increasing the quantity does not bring a noticeable improvements.}
	\vspace*{-1.5em}
	\label{table:scaling}
\end{table}
\begin{table}[t]
	\centering
	\renewcommand\arraystretch{1.1}
	\resizebox{0.75\linewidth}{!}{
		\begin{tabular}{c|ccc}
			\hline
			Fine-tune              & \begin{tabular}[c]{@{}c@{}}ReID \\ ($Rank1$) \end{tabular} & \begin{tabular}[c]{@{}c@{}}PAR \\ ($mA$) \end{tabular}  & \begin{tabular}[c]{@{}c@{}}HP \\ ($mIoU$) \end{tabular}\\ \hline
			25\% 	& 91.2          &   80.6 & 51.7 \\ 
			50\%   & 93.6        &   80.8  & 52.0  \\ 
			75\%  & 93.6  &    81.1  &  52.3  \\ 
			\hline  
			Full   &   93.6   & 80.7 & 52.3  \\
			\hline
		\end{tabular}
	}
	\caption{Downstream finetuning. SAIP achieves optimal results with 50\% finetuning epochs.}
	\vspace*{-1.5em}
	\label{tab:schedules}
\end{table}

\subsection{Main Results}
In this section, we compare SAIP with existing SSP methods and supervised pretraining methods across a wide range of HVP tasks. For comprehensive evaluations, we investigate three lightweight architectures, i.e., ViT\_Tiny~\cite{arch_vit_vit}, StarNet\_s3~\cite{arch_cnn_starnet} and TinyVit\_5M~\cite{arch_hybrid_tinyvit}.

\noindent\textbf{Single-person Discrimination.} As indicated in Table~\ref{table:sota_id_tasks}, SAIP achieves substantial performance improvement over existing methods. Using the ViT\_Tiny as the backbone, SAIP outperforms previous best method SOLIDER by 2.0\% and 6.4\% on Market and MSMT17, 3.7\% and 5.4\% on CUHK-PEDES and ICFG-PEDES, 2.1\% and 2.0\% on PA100K and PETAzs, respectively. Notably, SAIP's performance even surpasses that of supervised pretraining methods on these tasks. Furthermore, using a pretrained large model (i.e., PATH~\cite{hc_ssp_humanbench}) as the expert encoder, the performance improvements become much higher, achieving 3\%-13\%. This strong downstream performance indicates that the proposed SAIP serves not only as a SSP method but also an effective knowledge distillation solution. Similar improvements are also observed when utilizing StarNet or TinyVit as the backbone.

\noindent\textbf{Single-person Dense Predictions.} As illustrated in Table~\ref{table:sota_dp_tasks}, we fine-tune the parameters of SAIP models across three dense predictions tasks. With the same parameter count and same model architecture, the SAIP models exhibit superior performance compared to the counterparts of state-of-the-art SSP methods and supervised pretraining methods. For example, the SAIP pretrained ViT\_Tiny surpasses previous methods by 1.6\%-3.9\% in $AP$ scores for pose estimation task, by 2.1\%-6.2\% in $AP$ scores for landmark detection task, and by 6.3\%-11.7\% in $mIoU$ scores for part segmentation task. More interestingly, in the context of dense prediction tasks, employing a pretrained large model as the expert encoder does not yield a significant improvement, suggesting that visual patterns learned by SAIP are comparable to those learned from a pretrained large model.

\noindent\textbf{Multi-person Understanding.} Table~\ref{table:sota_mpvu_tasks} presents results for tasks that require to simultaneously solve person location, shape understanding and multi-granularity reasoning. Overall, SAIP surpasses previous SSP methods and supervised pretraining methods under the same setting. For example, SAIP family outperforms previous SSP methods by 1.0\%-3.8\% $AP$ scores for pedestrian detection, 2.3\%-5.1\% $mIoU$ scores for multiple human parsing, and 1.3\%-6.0\% $AP^{box}_{IoU+F_{1}}$ scores for part-level attribute parsing.

Based on these comparative experiments, one can conclude that proposed SAIP performs a general improvement across various HVP tasks and exhibits strong generalization.

\begin{figure}[t]
	\centering
	\includegraphics[width=0.99\linewidth]{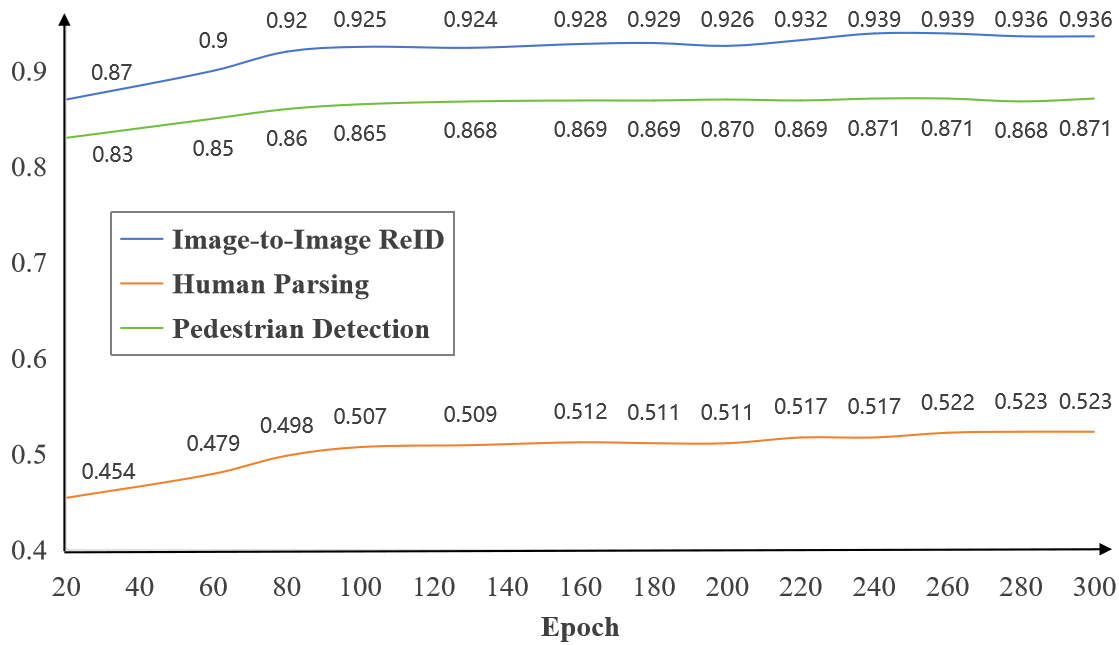}
	\caption{Training schedules. A longer training schedule does not give a noticeable improvement.}
	\vspace*{-1.5em}
	\label{fig:schedules}
\end{figure}
\subsection{Ablation Study}
In this section, we conduct ablation studies to investigate the major properties of SAIP. For fair comparison, we adopt the default settings, i.e., ViT\_Tiny serving as the backbone model and LUP1M adopted as pretraining dataset.  

\noindent \textbf{Effectiveness of the Devised Components in SAIP.} The SAIP is designed with three learning objectives, i.e., cross-scale matching (CSM), cross-scale reconstruction (CSR) and cross-scale search (CSS). We conduct a component-wise analysis by progressively using each objective to pretrain lightweight models, seeking to reveal the impact of each learning objective. Randomly initialized ViT\_Tiny (i.e., without pretraining) and its IN1K pretrained counterpart are used as baseline models. As reported in Table~\ref{table:abla_saip_component}, the pretrained ViT\_Tiny exhibits significantly superior performance compared to the randomly initialized variant, indicating the critical importance of pretraining. Furthermore, each cross-scale learning objective makes a positive contribution to performance, achieving results that are either superior or comparable to those of IN1K pretrained model. This indicates that a proper self-supervised learning strategy can yield better visual patterns than those derived from supervised learning. Additionally, it is evident that CSM, CSR, and CSS complement one another, collectively improving performance on downstream tasks. This illustrates that multi-task self-supervised learning is particularly beneficial for unlocking lightweight model's potential to learn general human visual patterns. 

\noindent \textbf{Impact of the Pretraining Dataset.} Existing works~\cite{hc_ssp_solider, hc_ssp_humanbench,  dataset_sapiens} emphasize the significance of the quantity of person images. In this work, we further extend the analysis by considering two additional dimensions of the pretraining dataset: the diversity of object categories and high fidelity of person images. To this end, we prepare five distinct datasets as listed in Table~\ref{table:scaling}. In addition to ImageNet-1K (IN1K)~\cite{dataset:in1k}, which contains 1000 object categories, we further utilize LUPerson dataset~\cite{dataset_luperson} to construct three datasets: LUP1M, LUP2M, and LUP4M with 1 million, 2 million, and 4 million person images, respectively. Notably, LUPerson contains significant noise. To develop a dataset with high fidelity person images, we collect 1 million images from 7 publicly available datasets (COCO, AIC, etc). For simplicity, the collected dataset is denoted as HP1M. 

Comparison results in Table~\ref{table:scaling} indicate that improving fidelity of person images benefits model performance more effectively than increasing diversity, while merely increasing the quantity does not bring a noticeable improvement. Furthermore, combining LUP1M with IN1K and HP1M leads to the best performance, demonstrating that high fidelity person images with a good diversity are particularly important for pretraining lightweight models.

\noindent\textbf{Training Efficiency in Downstream Tasks.} To investigate the impact of SAIP on downstream fine-tuning, we fine-tune SAIP models using four distinct training schedules. As shown in Table~\ref{tab:schedules}, the optimal performance is attained using 50\% of the fine-tuning epochs, presenting a rapid convergence. This indicates that SAIP is beneficial for accelerating convergence in fine-tuning stage, thereby establishing it as a computationally sustainable solution.

\noindent\textbf{Impact of Pretraining Epochs.} In general, large-scale pretraining often brings heavy training burden, which significantly limit affordability of SSP methods. However, we surprisingly find that proposed SAIP is beneficial for reducing training costs in pretraining stage. As shown in Fig~\ref{fig:schedules}, SAIP achieves optimal downstream performance at early stage of pretraining schedule, i.e., at 200-th epoch rather than 300-th epoch.

\begin{table}[t]
	\centering
	\renewcommand\arraystretch{1.3}
	\resizebox{0.95\linewidth}{!}{
		\begin{tabular}{c|c|c|c}
			\hline
			Method & Humanart & Chimpact-Pose & AP-10K \\ \hline
			DINO~\cite{ssp_dino}             & 65.7                 & 16.1                        & 58.0                \\ \hline
			MAE~\cite{ssp_mae}           & 65.1                 & 13.9                        & 48.8                \\ \hline
			MAE + DINO~\cite{ssp_mae_dino}     & 67.4                 & 18.2                       & 59.5                \\ \hline
			HAP~\cite{hc_ssp_hap}          & 66.0                & 13.2                       & 50.5                \\ \hline
			SOLIDER~\cite{hc_ssp_solider}          & 66.7                 & 16.0                        & 57.4                \\ \hline
			SAIP (Ours)             & \textbf{67.5}                 & \textbf{18.2}                        & \textbf{60.8}                \\ \hline
		\end{tabular}
	}
	\caption{Investigating generalization capability of HVP models by testing them on unseen domains, including: (1) person images with unseen styles, (2) animal images, and (3) common object images.}
	\label{tab:cross-domain}
\end{table}
\noindent\textbf{Cross-domain Generalizability of SAIP.} We examine cross-domain generalization capabilities using leave-one-domain-out evaluation, testing how self-supervised pretrained models transfer to novel domains absent from pretraining data. In our experimental design, we employ natural person images for pretraining and evaluate domain generalization on three distinct unseen domains: 1) stylized person images with various styles such as cartoons and sketches; 2) natural animal images, and 3) natural common object images. We conduct this investigation using three representative datasets, i.e., Humanart~\cite{dataset:humanart}, Chimpact-Pose~\cite{dataset:chimpact} and AP-10K~\cite{dataset:ap-10k}. Testing results presented in Tab~\ref{tab:cross-domain} indicate that our approach surpasses prior SSP methods. This demonstrates that the SAIP pretrained model learns general representations that are robust to domain changes and exhibits strong cross-domain generalization capability.

\begin{figure*}[t]
	\centering
	\includegraphics[width=0.75\linewidth]{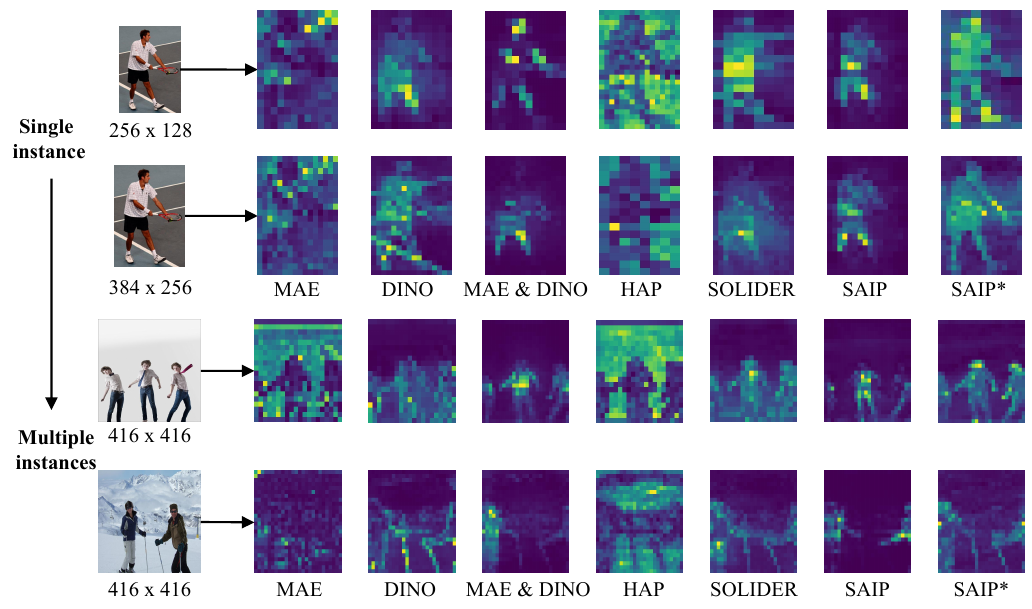}
	\vspace*{-0.5em}
	\caption{Qualitative comparison between proposed SAIP and state-of-the-art SSP methods.}
	\vspace*{-1.5em}
	\label{figure:vis_comp1}
\end{figure*}

\noindent\textbf{Qualitative Comparison.} We seek to reveal the learned visual patterns by conducting a comparative visualization analysis across different SSP methods. Specifically, we examine two scenarios: single instance across two scales and multiple instances within a single image. The visual patterns learned by various SSP methods are presented in Fig.~\ref{figure:vis_comp1}. Based on the visualization comparison, we have the following observations: First, most vision models, particularly those based on MIM (i.e., MAE and HAP), fail to capture visual patterns in both two scenarios. Second, SAIP-pretrained model demonstrates the ability to capture visual pattern for a person across multiple scales, but exhibit limitations in handling multiple instances. Third, when pretrained with a stronger expert encoder, SAIP-pretrained model effectively captures meaningful patterns in both scenarios. These observations, corroborated by quantitative comparisons, reveal a clear correlation: vision models that capture more diverse human visual patterns consistently achieve superior downstream performance.

\section{Conclusion}
In this paper, we identify two key limitations in adapting existing HVP models to real-world applications, i.e., specific visual patterns and excessively large model size. To handle these, we propose SAIP, the first self-supervised pretraining framework designed to establish lightweight and generalizable models for HVP. Extensive experiments on a wide range of HVP tasks demonstrate the effectiveness and strong generalization capabilities of SAIP. Furthermore, we highlight two useful findings related to pretraining lightweight vision models. These findings suggest that if a proper training objective is explored, lightweight architectures can serve as a good generalizable model for HVP. We hope this work can bring inspiration to the community.

\section*{A. SAIP Pretraining Details}
We first present the implementation details of major components of SAIP, including the architectures of tasks decoders, pretraining settings and pretraining datasets. 

\noindent \textbf{CSM decoder} consists of a Multi-Layer Perceptron (MLP) block, a $l_2$ normalization layer and a projection layer, following the design principles of DINO~\cite{ssp_dino}. The MLP block, which comprises three non-linear layers, transforms image tokens of $D$ dimensions (i.e., 192) into a latent representation of 256 dimensions. Subsequently, the latent representation is normalized by $l_2$ normalization layer and further mapped to a high-dimensional instance representation (i.e., 65536) by the projection layer.

\noindent \textbf{CSR decoder} is a stack of 6 transformer blocks, each incorporating a multi-head attention block, LayerNorm layers applied before and after attention block, and nonlinear layers. The CSR decoder processes sequential tokens encoded by the image encoder, including tokens from the anchor image and visible tokens from the masked image. 

\noindent \textbf{CSS decoder} includes two attention blocks identical to those in the CSR decoder. In addition, a cross-attention block, four convolution layers and one predictive linear layer are appended to the attention blocks. Specifically, the convolution layers and the predictive linear layer are used for binary mask extraction. The binary mask represents a rectangular region corresponding to an instance.

\noindent \textbf{Pretraining setting.} The default pretraining settings are presented in Table ~\ref{table:pretraining-setting}. All transformer blocks are initialized using the Xavier uniform. The learning rate is linearly warmed up over the first 10 epochs and subsequently decayed using a cosine schedule. The anchor image size is set to $256\times128$, while the multi-person image has a size of $224\times224$.
\begin{table}[t]
	\centering
	\renewcommand\arraystretch{1.1}
	\resizebox{0.8\linewidth}{!}{
		\begin{tabular}{c|c}
			\hline
			\multicolumn{1}{c}{\textbf{Config}}                 & \multicolumn{1}{c}{\textbf{Value}}               \\ \hline
			Optimizer                       & AdamW           \\
			Base Learning Rate              & 2.5e-4                       \\
			Weight Decay                    & 0.05                         \\
			Optimizer Momentum              & $\beta_1, \beta_2 = 0.9, 0.95$\\
			Batch Size per gpu              & 256                         \\
			Learning Rate Schedule          & Cosine scheduler   \\
			Warmup Epochs                   & 10                           \\
			\hline
		\end{tabular}
	}
	\caption{Pretraining setting.}
	\vspace*{-2em}
	\label{table:pretraining-setting}
\end{table}

\noindent \textbf{Pretraining dataset.} In the paper, we investigate the influence of pretraining data by constructing 5 different person-centric datasets, i.e., IN1K, LUP1M, LUP2M, LUP4M and HP1M. Specifically, LUPxM is derived from LUPerson dataset~\cite{dataset_luperson}, and HP1M is a high-quality dataset collected from 7 publicly available datasets, i.e., AIC~\cite{dataset:pose-aic}, COCO~\cite{dataset:coco}, CIHP~\cite{dataset:cihp}, MHPv2~\cite{dataset:mhp}, CrowdHuman~\cite{dataset:crowdhuman}, Fashionpedia~\cite{dataset:fashionpedia}, and HumanArt~\cite{dataset:humanart}. Notably, these datasets provide manually annotated bounding boxes. Therefore, we collect images from the training sets of these 7 datasets, and use the provided bounding boxes to crop person instances from collected images, resulting in approximately 1 million high-fidelity person images. Detailed statistics of HP1M are presented in Table~\ref{table:hp1m}.

\begin{table}[t]
	\centering
	\renewcommand\arraystretch{1.2}
	\resizebox{0.9\linewidth}{!}{
		\begin{tabular}{p{3cm}|>{\centering\arraybackslash}m{3cm}}
			\hline
			\multicolumn{1}{l|}{\textbf{dataset}}                 & \multicolumn{1}{c}{\textbf{\#images}}               \\ \hline
			AIC~\cite{dataset:pose-aic}                       & 378352           \\
			COCO~\cite{dataset:coco}              & 149813                       \\
			CIHP~\cite{dataset:cihp}                    & 93207                         \\
			MHPv2~\cite{dataset:mhp}              & 41056 \\
			CrowdHuman~\cite{dataset:crowdhuman}              & 428109                         \\
			Fashionpedia~\cite{dataset:fashionpedia}          & 45623   \\
			HumanArt~\cite{dataset:humanart}                 & 81976                           \\
			\hline
			Total   &   1218136 \\
			\hline
		\end{tabular}
	}
	\caption{The statistics of HP1M.}
	\vspace*{-2em}
	\label{table:hp1m}
\end{table}

\section*{B. Downstream Fine-tuning Details}
For the evaluation of pretrained methods, we adopt representative solutions in downstream tasks as baselines, replacing their backbones with the pretrained models.

\noindent\textbf{I2I person ReID.} We use the TransReID~\cite{ds_reid_transreid} implemented in~\cite{hc_ssp_solider} as the basic solution for I2I ReID. All models are fine-tuned on Market1501~\cite{dataset:market_reid} \texttt{train} and MSMT17~\cite{dataset:msmt_reid} \texttt{train} respectively, and further evaluated on Market1501 \texttt{test} and MSMT17 \texttt{test}. The training process consists of 120 epochs with a default learning rate of 2e-4. During both training phase and testing phase, the image size is set to $256\times128$.

\noindent\textbf{T2I person ReID.} We use the IRRA~\cite{ds_t2ireid_irra} with its publicly available code base to evaluate pretrained models. All pretrained models are fine-tuned on CUHK-PEDES~\cite{dataset:t2i_reid_cuhk} \texttt{train} or ICFG-PEDES~\cite{dataset:t2i_reid_icfg} \texttt{train}, and tested on CUHK-PEDES \texttt{test} or ICFG-PEDES \texttt{test}. Each model is fine-tuned for 60 epochs, with the image size during both training phase and testing phase set to $384\times128$ by default. To accelerate convergence, we adopt relatively large batch size (i.e., 128) and relatively large learning rate (i.e., 1e-4), which are different from original settings used in~\cite{ds_t2ireid_irra} (i.e., 64 and 1e-5).

\noindent\textbf{Person attribute recognition.} We use the PAR~\cite{dataset:pa100k}as the baseline model, as implemented in~\cite{hc_ssp_solider}. The model is initialized with pretrained models and further fine-tuned on PA100~\cite{dataset:pa100k} and PETAzs~\cite{dataset:peta} respectively. The image size is set to $256\times128$, and all models are fine-tuned for 25 epochs by default.

\noindent\textbf{Human pose estimation.} We use the ViTPose~\cite{ds_pose_vitpose} as the basic solution for HPE, implemented with open-source mmpose library\footnote{\url{https://github.com/open-mmlab/mmpose}}. The ViTPose model is initialized with pretrained models, fine-tuned on COCO keypoint~\cite{dataset:coco} \texttt{train2017}, and evaluated on COCO keypoint \texttt{val2017}. In our experiments, the image size is set to $256\times192$ and other settings follow the default configuration in mmpose.

\noindent\textbf{Person landmarks detection.} We use the mmpose to implement PLD model based on ViTPose, which is fine-tuned on whole-body COCO~\cite{dataset:wholebody-coco} \texttt{train} and evaluated on whole-body COCO \texttt{val}. The image size is set to $256\times192$ in our experiments and other settings follow the default configuration in mmpose.

\noindent\textbf{Body part segmentation.} We implement body part segmentation model based on SCHP~\cite{ds_hp_schp} method, and use the open-sourced code provided in~\cite{hc_ssp_solider} for training and testing. All pretrained models are fine-tuned on LIP~\cite{dataset:LIP} dataset and evaluated on LIP test set. Additionally, we adopt AdamW~\cite{optim_adamw} as the optimizer during fine-tuning. Other implementation settings follow those of SCHP~\cite{ds_hp_schp}.

\noindent\textbf{Pedestrian detection.} We adopt CrowdDet~\cite{ds_pdetection_crowddet} with pretrained backbone models as the basic solution for pedestrian detection. Those models are fine-tuned on CrowdHuman~\cite{dataset:crowdhuman} using open-sourced mmdetection\footnote{\url{https://github.com/open-mmlab/mmdetection}}. We use the default configuration provided in mmdetection for training and testing, where AdamW optimizer with a learning rate of 2e-4 are used for training.

\noindent\textbf{Multiple human parsing.} We evaluate pretrained models on MHP using the Parsing-RCNN~\cite{ds_hp_parsingrcnn} implemented in~\cite{ds_mhp_cpiparser}. The pretrained models are fine-tuned on CIHP~\cite{dataset:cihp} \texttt{train} for 25 training epochs (i.e., $1\times$ schedule), and evaluated on CIHP \texttt{val}.

\noindent\textbf{Part-level attribute parsing.} We evaluate pretrained models on PAP using the KE-RCNN~\cite{ds_part_parse_kercnn}. The pretrained models are fine-tuned on Fashionpedia~\cite{dataset:fashionpedia} \texttt{train} for 32 training epochs (i.e., $1\times$ schedule), and evaluated on Fashionpedia \texttt{val}.

\bibliographystyle{IEEEtran}
\bibliography{main}

\begin{thebibliography}{10}
\providecommand{\url}[1]{#1}
\csname url@samestyle\endcsname
\providecommand{\newblock}{\relax}
\providecommand{\bibinfo}[2]{#2}
\providecommand{\BIBentrySTDinterwordspacing}{\spaceskip=0pt\relax}
\providecommand{\BIBentryALTinterwordstretchfactor}{4}
\providecommand{\BIBentryALTinterwordspacing}{\spaceskip=\fontdimen2\font plus
\BIBentryALTinterwordstretchfactor\fontdimen3\font minus
  \fontdimen4\font\relax}
\providecommand{\BIBforeignlanguage}[2]{{%
\expandafter\ifx\csname l@#1\endcsname\relax
\typeout{** WARNING: IEEEtran.bst: No hyphenation pattern has been}%
\typeout{** loaded for the language `#1'. Using the pattern for}%
\typeout{** the default language instead.}%
\else
\language=\csname l@#1\endcsname
\fi
#2}}
\providecommand{\BIBdecl}{\relax}
\BIBdecl

\bibitem{ssp_dino}
M.~Caron, H.~Touvron, I.~Misra, H.~J\'egou, J.~Mairal, P.~Bojanowski, and
  A.~Joulin, ``Emerging properties in self-supervised vision transformers,'' in
  \emph{ICCV}, 2021, pp. 9630--9640.

\bibitem{ssp_mae}
K.~He, X.~Chen, S.~Xie, Y.~Li, P.~Doll\'ar, and R.~Girshick, ``Masked
  autoencoders are scalable vision learners,'' in \emph{CVPR}, June 2022, pp.
  16\,000--16\,009.

\bibitem{ssp_eva}
Y.~Fang, W.~Wang, B.~Xie, Q.~Sun, L.~Wu, X.~Wang, T.~Huang, X.~Wang, and
  Y.~Cao, ``Eva: Exploring the limits of masked visual representation learning
  at scale,'' in \emph{CVPR}, 2023, pp. 19\,358--19\,369.

\bibitem{ssp_mae-lite}
S.~Wang, J.~Gao, Z.~Li, X.~Zhang, and W.~Hu, ``A closer look at self-supervised
  lightweight vision transformers,'' 2023.

\bibitem{ssp_mae_dino}
N.~Park, W.~Kim, B.~Heo, T.~Kim, and S.~Yun, ``What do self-supervised vision
  transformers learn?'' in \emph{ICLR}, 2023.

\bibitem{ssp_maefuse}
J.~Li, J.~Jiang, P.~Liang, J.~Ma, and L.~Nie, ``Maefuse: Transferring omni
  features with pretrained masked autoencoders for infrared and visible image
  fusion via guided training,'' \emph{IEEE Transactions on Image Processing},
  vol.~34, pp. 1340--1353, 2025.

\bibitem{ssp_tinymim}
S.~Ren, F.~Wei, Z.~Zhang, and H.~Hu, ``Tinymim: An empirical study of
  distilling mim pre-trained models,'' in \emph{CVPR}, 2023, pp. 3687--3697.

\bibitem{hc_ssp_hap}
J.~Yuan, X.~Zhang, H.~Zhou, J.~Wang, Z.~Qiu, Z.~Shao, S.~Zhang, S.~Long,
  K.~Kuang, K.~Yao, J.~Han, E.~Ding, L.~Lin, F.~Wu, and J.~Wang, ``Hap:
  Structure-aware masked image modeling for human-centric perception,'' in
  \emph{NeurIPS}, vol.~36, 2023, pp. 50\,597--50\,616.

\bibitem{hc_ssp_hcmoco}
F.~Hong, L.~Pan, Z.~Cai, and Z.~Liu, ``Versatile multi-modal pre-training for
  human-centric perception,'' in \emph{CVPR}, 2022, pp. 16\,156--16\,166.

\bibitem{hc_ssp_humanbench}
S.~Tang, C.~Chen, Q.~Xie, M.~Chen, Y.~Wang, Y.~Ci, L.~Bai, F.~Zhu, H.~Yang,
  L.~Yi, R.~Zhao, and W.~Ouyang, ``Humanbench: Towards general human-centric
  perception with projector assisted pretraining,'' in \emph{CVPR}, 2023, pp.
  21\,970--21\,982.

\bibitem{hc_ssp_solider}
W.~Chen, X.~Xu, J.~Jia, H.~Luo, Y.~Wang, F.~Wang, R.~Jin, and X.~Sun, ``Beyond
  appearance: A semantic controllable self-supervised learning framework for
  human-centric visual tasks,'' in \emph{CVPR}, 2023, pp. 15\,050--15\,061.

\bibitem{hc_ssp_unihcp}
Y.~Ci, Y.~Wang, M.~Chen, S.~Tang, L.~Bai, F.~Zhu, R.~Zhao, F.~Yu, D.~Qi, and
  W.~Ouyang, ``Unihcp: A unified model for human-centric perceptions,'' in
  \emph{CVPR}, 2023, pp. 17\,840--17\,852.

\bibitem{dataset_sapiens}
R.~Khirodkar, T.~Bagautdinov, J.~Martinez, S.~Zhaoen, A.~James, P.~Selednik,
  S.~Anderson, and S.~Saito, ``Sapiens: Foundation for human vision models,''
  2024.

\bibitem{optim_scaling_laws}
J.~Kaplan, S.~McCandlish, T.~Henighan, T.~B. Brown, B.~Chess, R.~Child,
  S.~Gray, A.~Radford, J.~Wu, and D.~Amodei, ``Scaling laws for neural language
  models,'' \emph{arXiv preprint arXiv:2001.08361}, 2020.

\bibitem{hc_ssp_hqnet}
S.~Jin, S.~Li, T.~Li, W.~Liu, C.~Qian, and P.~Luo, ``You only learn one query:
  Learning unified human query for single-stage multi-person multi-task
  human-centric perception,'' \emph{arXiv preprint arXiv:2312.05525}, 2024.

\bibitem{dataset_luperson}
D.~Fu, D.~Chen, J.~Bao, H.~Yang, L.~Yuan, L.~Zhang, H.~Li, and D.~Chen,
  ``Unsupervised pre-training for person re-identification,'' pp.
  14\,745--14\,754, 2021.

\bibitem{ds_attr_rec_par}
J.~Jia, X.~Chen, and K.~Huang, ``Spatial and semantic consistency
  regularizations for pedestrian attribute recognition,'' in \emph{ICCV}, 2021,
  pp. 962--971.

\bibitem{ds_reid_transreid}
S.~He, H.~Luo, P.~Wang, F.~Wang, H.~Li, and W.~Jiang, ``Transreid:
  Transformer-based object re-identification,'' in \emph{ICCV}, 2021, pp.
  15\,013--15\,022.

\bibitem{ds_t2ireid_irra}
D.~Jiang and M.~Ye, ``Cross-modal implicit relation reasoning and aligning for
  text-to-image person retrieval,'' in \emph{CVPR}, 2023.

\bibitem{ds_pose_vitpose}
Y.~Xu, J.~Zhang, Q.~Zhang, and D.~Tao, ``Vi{TP}ose: Simple vision transformer
  baselines for human pose estimation,'' in \emph{NeurIPS}, 2022.

\bibitem{ds_hp_parsingrcnn}
L.~Yang, Q.~Song, Z.~Wang, and M.~Jiang, ``Parsing {R-CNN} for instance-level
  human analysis,'' in \emph{CVPR}, 2019.

\bibitem{ds_hp_schp}
P.~Li, Y.~Xu, Y.~Wei, and Y.~Yang, ``Self-correction for human parsing,''
  \emph{IEEE Transactions on Pattern Analysis and Machine Intelligence},
  vol.~44, no.~6, pp. 3260--3271, 2022.

\bibitem{dataset:LIP}
X.~Liang, K.~Gong, X.~Shen, and L.~Lin, ``Look into person: Joint body parsing
  \& pose estimation network and a new benchmark,'' \emph{IEEE Transactions on
  Pattern Analysis and Machine Intelligence}, vol.~41, no.~4, pp. 871--885,
  2019.

\bibitem{ds_part_parse_kercnn}
X.~Wang, J.~Song, X.~Chen, L.~Cheng, L.~Gao, and H.~T. Shen, ``Ke-rcnn:
  Unifying knowledge-based reasoning into part-level attribute parsing,''
  \emph{IEEE Transactions on Cybernetics}, vol.~53, no.~11, pp. 7263--7274,
  2023.

\bibitem{ds_pdetection_crowddet}
X.~Chu, A.~Zheng, X.~Zhang, and J.~Sun, ``Detection in crowded scenes: One
  proposal, multiple predictions,'' in \emph{CVPR}, 2020.

\bibitem{dataset:cihp}
K.~Gong, X.~Liang, Y.~Li, Y.~Chen, M.~Yang, and L.~Lin, ``Instance-level human
  parsing via part grouping network,'' in \emph{ECCV}, 2018, pp. 805--822.

\bibitem{dataset:fashionpedia}
M.~Jia, M.~Shi, M.~Sirotenko, Y.~Cui, C.~Cardie, B.~Hariharan, H.~Adam, and
  S.~Belongie, ``Fashionpedia: Ontology, segmentation, and an attribute
  localization dataset,'' in \emph{ECCV}, 2020.

\bibitem{ssp_moco}
K.~He, H.~Fan, Y.~Wu, S.~Xie, and R.~Girshick, ``Momentum contrast for
  unsupervised visual representation learning,'' in \emph{CVPR}, 2020, pp.
  9726--9735.

\bibitem{ssp_oadp}
Y.~Zhang, T.~Zhang, H.~Zhu, Z.~Chen, S.~Mi, X.~Peng, and X.~Geng, ``Object
  adaptive self-supervised dense visual pre-training,'' \emph{IEEE Transactions
  on Image Processing}, vol.~34, pp. 2228--2240, 2025.

\bibitem{arch_cnn_starnet}
X.~Ma, X.~Dai, Y.~Bai, Y.~Wang, and Y.~Fu, ``Rewrite the stars,'' in
  \emph{CVPR}, 2024.

\bibitem{arch_cnn_cspnext}
C.-Y. Wang, H.-Y.~M. Liao, I.-H. Yeh, Y.-H. Wu, P.-Y. Chen, and J.-W. Hsieh,
  ``Cspnet: A new backbone that can enhance learning capability of cnn,''
  \emph{arXiv preprint arXiv:1911.11929}, 2019.

\bibitem{arch_vit_vit}
A.~Dosovitskiy, L.~Beyer, A.~Kolesnikov, D.~Weissenborn, X.~Zhai,
  T.~Unterthiner, M.~Dehghani, M.~Minderer, G.~Heigold, S.~Gelly, J.~Uszkoreit,
  and N.~Houlsby, ``An image is worth 16x16 words: Transformers for image
  recognition at scale,'' in \emph{ICLR}, 2021.

\bibitem{arch_hybrid_edgenext}
M.~Maaz, A.~Shaker, H.~Cholakkal, S.~Khan, S.~W. Zamir, R.~M. Anwer, and F.~S.
  Khan, ``Edgenext: Efficiently amalgamated cnn-transformer architecture for
  mobile vision applications,'' 2022.

\bibitem{arch_hybrid_tinyvit}
K.~Wu, J.~Zhang, H.~Peng, M.~Liu, B.~Xiao, J.~Fu, and L.~Yuan, ``Tinyvit: Fast
  pretraining distillation for small vision transformers,'' in \emph{ECCV},
  2022.

\bibitem{data_aug_copypaste}
G.~Ghiasi, Y.~Cui, A.~Srinivas, R.~Qian, T.-Y. Lin, E.~D. Cubuk, Q.~V. Le, and
  B.~Zoph, ``Simple copy-paste is a strong data augmentation method for
  instance segmentation,'' in \emph{CVPR}, 2021, pp. 2917--2927.

\bibitem{dataset:market_reid}
L.~Zheng, L.~Shen, L.~Tian, S.~Wang, J.~Wang, and Q.~Tian, ``Scalable person
  re-identification: A benchmark,'' in \emph{ICCV}, 2015.

\bibitem{dataset:msmt_reid}
L.~Wei, S.~Zhang, W.~Gao, and Q.~Tian, ``Person transfer gan to bridge domain
  gap for person re-identification,'' in \emph{CVPR}, 2018, pp. 79--88.

\bibitem{dataset:t2i_reid_cuhk}
S.~Li, T.~Xiao, H.~Li, B.~Zhou, D.~Yue, and X.~Wang, ``Person search with
  natural language description,'' \emph{arXiv preprint arXiv:1702.05729}, 2017.

\bibitem{dataset:t2i_reid_icfg}
Z.~Ding, C.~Ding, Z.~Shao, and D.~Tao, ``Semantically self-aligned network for
  text-to-image part-aware person re-identification,'' \emph{arXiv preprint
  arXiv:2107.12666}, 2021.

\bibitem{dataset:pa100k}
X.~Liu, H.~Zhao, M.~Tian, L.~Sheng, J.~Shao, J.~Yan, and X.~Wang,
  ``Hydraplus-net: Attentive deep features for pedestrian analysis,'' in
  \emph{Proceedings of the IEEE international conference on computer vision},
  2017, pp. 1--9.

\bibitem{dataset:peta}
Y.~DENG, P.~Luo, C.~C. Loy, and X.~Tang, ``Pedestrian attribute recognition at
  far distance,'' 2014, p. 789–792.

\bibitem{dataset:coco}
T.~Lin, M.~Maire, S.~J. Belongie, J.~Hays, P.~Perona, D.~Ramanan,
  P.~Doll{\'{a}}r, and C.~L. Zitnick, ``Microsoft {COCO:} common objects in
  context,'' in \emph{ECCV}, 2014.

\bibitem{dataset:wholebody-coco}
S.~Jin, L.~Xu, J.~Xu, C.~Wang, W.~Liu, C.~Qian, W.~Ouyang, and P.~Luo,
  ``Whole-body human pose estimation in the wild,'' in \emph{ECCV}, 2020.

\bibitem{dataset:crowdhuman}
S.~Shao, Z.~Zhao, B.~Li, T.~Xiao, G.~Yu, X.~Zhang, and J.~Sun, ``Crowdhuman: A
  benchmark for detecting human in a crowd,'' \emph{arXiv preprint
  arXiv:1805.00123}, 2018.

\bibitem{optim_adamw}
I.~Loshchilov and F.~Hutter, ``Decoupled weight decay regularization,''
  \emph{arXiv preprint arXiv:1711.05101}, 2017.

\bibitem{optim_cosine_anneal}
------, ``Sgdr: Stochastic gradient descent with warm restarts,'' \emph{arXiv
  preprint arXiv:1608.03983}, 2016.

\bibitem{arch_cnn_resnet}
K.~He, X.~Zhang, S.~Ren, and J.~Sun, ``Deep residual learning for image
  recognition,'' in \emph{CVPR}, 2016.

\bibitem{dataset:in1k}
J.~Deng, W.~Dong, R.~Socher, L.-J. Li, K.~Li, and L.~Fei-Fei, ``Imagenet: A
  large-scale hierarchical image database,'' in \emph{CVPR}, 2009, pp.
  248--255.

\bibitem{dataset:humanart}
X.~Ju, A.~Zeng, J.~Wang, Q.~Xu, and L.~Zhang, ``Human-art: A versatile
  human-centric dataset bridging natural and artificial scenes,'' in
  \emph{Proceedings of the IEEE/CVF Conference on Computer Vision and Pattern
  Recognition}, 2023.

\bibitem{dataset:chimpact}
X.~Ma, S.~Kaufhold, J.~Su, W.~Zhu, J.~Terwilliger, A.~Meza, Y.~Zhu, F.~Rossano,
  and Y.~Wang, ``Chimpact: A longitudinal dataset for understanding chimpanzee
  behaviors,'' \emph{Advances in Neural Information Processing Systems},
  vol.~36, pp. 27\,501--27\,531, 2023.

\bibitem{dataset:ap-10k}
H.~Yu, Y.~Xu, J.~Zhang, W.~Zhao, Z.~Guan, and D.~Tao, ``Ap-10k: A benchmark for
  animal pose estimation in the wild,'' in \emph{Advances in Neural Information
  Processing Systems}, 2021.

\bibitem{dataset:pose-aic}
J.~Wu, H.~Zheng, B.~Zhao, Y.~Li, B.~Yan, R.~Liang, W.~Wang, S.~Zhou, G.~Lin,
  Y.~Fu \emph{et~al.}, ``Ai challenger: A large-scale dataset for going deeper
  in image understanding,'' \emph{arXiv preprint arXiv:1711.06475}, 2017.

\bibitem{dataset:mhp}
J.~Zhao, J.~Li, Y.~Cheng, T.~Sim, S.~Yan, and J.~Feng, ``Understanding humans
  in crowded scenes: Deep nested adversarial learning and {A} new benchmark for
  multi-human parsing,'' in \emph{ACM MM}, 2018.

\bibitem{ds_mhp_cpiparser}
X.~Wang, X.~Chen, L.~Gao, J.~Song, and H.~T. Shen, ``Cpi-parser: Integrating
  causal properties into multiple human parsing,'' \emph{IEEE Transactions on
  Image Processing}, vol.~33, pp. 5771--5782, 2024.

\end{thebibliography}
\vfill

\end{document}